\documentclass{l4dc2027}
\title[From Self-Play RL to Constrained Control]{From Self-Play RL to Constrained Control: Distilling DiffMPC Students with DAgger}
\usepackage{times}
\usepackage{amssymb}
\usepackage{wrapfig}
\ifodd 0

\newcommand{\com}[1]{\textbf{\color{red}(Parinaz: #1)}}
\newcommand{\coml}[1]{\textbf{\color{orange}(Luca: #1)}}
\else

\newcommand{\com}[1]{}
\newcommand{\coml}[1]{}
\fi

\author{%
 \Name{Luca M. Hartmann} \Email{lhartmann@ucsd.edu}\\
 \addr University of California San Diego
 \AND
 \Name{Parinaz Naghizadeh} \Email{parinaz@ucsd.edu}\\
 \addr University of California San Diego%
}

\begin{document}

\maketitle

\begin{abstract}%
 Self-play reinforcement learning (RL) produces strong driving policies, but these policies are hard to verify or interpret. Differentiable model predictive control (DiffMPC) can enforce explicit constraints, but it is usually trained by imitating logged data or with costly RL. To add constraints to an already trained policy, we keep a PPO teacher's backbone and replace its action head with a DiffMPC that enforces road boundary half-planes and agent ellipses. We then distill the student with DAgger in the teacher's self-play simulator. Over 9 seeds, the constrained student of a smaller teacher matches its teacher on nuPlan and beats it on CARLA and on out-of-distribution Waymo scenes. For a stronger teacher, the student has fewer failures on nuPlan and Waymo, while staying level on CARLA. Its distillation takes no more than 30\% of the teacher's training time.
\end{abstract}

\begin{keywords}%
  Differentiable model predictive control, policy distillation, DAgger, self-play reinforcement learning, autonomous driving, safety constraints%
\end{keywords}

\section{Introduction}
Machine learning for robotics has made fast progress in recent years~\citep{tang2025drlrobotics,chen2024e2esurvey}. While earlier work mostly relied on imitation learning (IL)~\citep{lemero2022ilsurvey}, reinforcement learning (RL) for training and post-training has gained traction~\citep{kiran2022drlad,lu2023imitation,kaufmann2023champion}. Fast simulators such as PufferDrive~\citep{cornelisse2026humanlike,rowe2026gigapixel} train neural driving policies over millions of kilometers in self-play and reach super-human performance~\citep{cusumanotowner2025gigaflow}. % GigaFlow
Vision-based end-to-end policies follow the same trend. They can be trained with RL only~\citep{yin2026pictura}, % Pictura
post-trained with RL~\citep{li2026worldengine}, % WorldEngine
or distilled from RL-trained policies~\citep{dacol2026opted}. % OPTED

However, a main limitation of such pure end-to-end policies is that they are hard to verify and interpret. In safety-critical applications, such as autonomous driving, safety during training, verifiability, and interpretability are all important~\citep{chen2024e2esurvey, brunke2022safe}. 
Motivated by these challenges, a line of work in robotics research has combined learning with classic optimal control methods, such as predictive safety filters (PSFs)~\citep{wabersich2018linear} and differentiable model predictive control (DiffMPC)~\citep{amos2018differentiable}. These methods enforce explicit, checkable constraints. When the constraints are computed independently of the policy, e.g., from the road and other agents, they add redundancy to the system and can help in scenes the policy has not seen. They also make decisions interpretable: a failure can be traced back to the constraints, the cost function, or the targets.

In this paper, our goal is to add safety constraints to an already trained, strong end-to-end policy without losing its performance. We propose to leverage DiffMPC for this goal, as it can be naturally integrated as the action head of an end-to-end policy. DiffMPC is trained jointly with the policy, so the policy learns with the constraints. So far, DiffMPC policies have mostly been trained with imitation learning or behavior cloning on demonstrations~\citep{huang2024dipp,karkus2023diffstack,gurumurthy2025deqmpc}. More recent work trains DiffMPCs with RL, but this is either expensive~\citep{romero2026acmpc} % AC-MPC
or has so far been demonstrated only in small navigation tasks with a single controlled robot and known obstacle motion~\citep{wu2026acsmpc}. % fast-trained AC-MPC 
In this work, we propose to \emph{distill} a DiffMPC student from an existing RL-trained end-to-end policy to overcome these limitations. \looseness=-1

We focus our study on autonomous driving, as it has a clear use case for verifiable constraints and well-developed simulators. Our approach, however, is not limited to autonomous driving and can be transferred to other robotics control and learning tasks. We train our policies in PufferDrive~\citep{rowe2026gigapixel} on maps and scenes from CARLA~\citep{dosovitskiy2017carla} and nuPlan~\citep{caesar2021nuplan}, and evaluate them on CARLA, nuPlan, and the Waymo Open Dataset~\citep{ettinger2021womd}.
Our main contributions are as follows:
\begin{itemize}
    \item We propose a pipeline that adds a DiffMPC layer to a trained end-to-end policy. Specifically, first, we train a neural teacher policy with PPO in self-play. Second, we keep its backbone and replace its action head with a constrained DiffMPC. Third, we distill the new architecture from the teacher with DAgger. To the best of our knowledge, this is the first large-scale distillation of a DiffMPC student from an end-to-end teacher in self-play.
    \item We propose a constraint formulation that makes this distillation work: fixed-sum cost weights with learned node weights, route-anchored road boundary constraints, and agent ellipses.
    \item We evaluate the pipeline on CARLA, nuPlan, and Waymo scenes with two teacher sizes and nine seeds per policy. The constrained students match or beat their teachers on every test set while adding explicit constraints, including on the out-of-distribution Waymo scenes, where they leave the road 16--26\% less often than their teachers. For the larger teacher, distillation adds no more than 30\% to its training time.
\end{itemize}

Our code, along with videos of driving scenarios, 
is available at~\url{https://github.com/lucahart/selfplay-to-mpc}.

\coml{ Mention MLP somewhere here and in e2e policy preliminaries.}

\section{Related work}\label{sec:related-work}

\citet{amos2018differentiable} made differentiable MPC scalable, and it has since gained traction in the robotics community~\citep{hewing2020learning,reiter2025synthesis}. In our application of autonomous driving, there is a broad literature on policies trained through IL, including~\cite{huang2024dipp, karkus2023diffstack}.\looseness=-1

DAgger distillation~\citep{ross2011dagger} improves other distillation methods, such as behavior cloning~\citep{bain1995behavioral-cloning}, by allowing the distilled student policy to drive itself.

DiffMPC RL training and imitation learning are being enabled by the latest advancements in DiffMPC solvers that can run on GPUs. \citet{adabag2025diffmpcgpu} showed large speedups for batched solving, and \citet{bravopalacios2026turbompc} scaled to long horizons. \cite{hose2026warpmpc} further improved solving performance for batches on the GPU, making it a perfect fit for large-scale training. 

The use of half-planes and ellipse constraints for MPC in autonomous driving has been covered in~\citep{schwarting2018safe, liniger2015optimization}. Recent DiffMPC setups with constraints include \citet{zhang2026smoothedturbompc} with track and actuator bounds, and \citet{rickenbach2025zipmpc, trisovic2026vigormpc} with track boundary constraints.

% Closely related work
Last, we discuss the literature most closely related to our work. 
\citet{rickenbach2025zipmpc} use imitation learning to train a short-horizon DiffMPC to match a long-horizon MPC. \citet{romero2026acmpc} train a DiffMPC policy with actor-critic RL, and report significant training costs. Closest to our constraint formulation, \citet{wu2026acsmpc} add half-space state constraints with bounded slack to an actor-critic DiffMPC and learn its cost parameters, demonstrated on navigation of a single controlled robot among moving obstacles with known motion. \citet{zhang2026smoothedturbompc} show a highly efficient RL trainer for DiffMPC, which relies on first-order RL and therefore needs a differentiable simulator and reward. \citet{papuc2026lmpg} use multi-layer perceptrons (MLPs) and a differentiable trajectory optimizer for two-drone racing, with collision avoidance only as a soft penalty. \cite{zarrouki2026sgrl} use RL to train the cost function of a nonlinear MPC in a single-agent simulation.

The above contributions have been central in pushing the frontier of DiffMPC research. This work is built upon and inspired by many of the mentioned papers. Unlike the above works, we (1) distill a DiffMPC student from an end-to-end RL teacher with DAgger, (2) train both in multi-agent autonomous-driving self-play, and (3) treat the constrained DiffMPC as a post-training step that adds redundancy to an existing policy.

\section{Preliminaries}\label{sec:preliminaries}

\subsection{End-to-end policies}

We refer to end-to-end policies as networks that have been trained as a whole, from raw data inputs to final output commands. A common end-to-end policy is a feed-forward neural network, also referred to as an MLP. Such networks have shown great popularity compared to policies with separately trained components~\citep{yurtsever2020survey}, as the communication between components does not need to be designed by hand. 

End-to-end policies commonly consist of three main components: (i) an encoder, that takes the observation and outputs a latent space representation; (ii) a backbone, that represents the learned heart of the policy; and (iii) a decoder, that turns the latent space into a trajectory or control actions.

There are three common ways to train an end-to-end policy: (1) Imitation learning (IL) trains a policy to imitate, e.g., human driving data. It requires no simulation and is trained with a fixed set of labels. This method suffers from poor generalizability and quickly fails in closed-loop simulation or real-world deployment~\citep{rowe2026gigapixel}. (2) Reinforcement learning (RL) uses a simulation, where a policy can explore different states and learn by its actions. It is often used as a post-training step to improve a policy that was pre-trained with IL~\citep{li2026worldengine}. RL, however, often requires many steps when training a policy, making it expensive in many scenarios~\citep{romero2026acmpc}. (3) Distillation uses a policy that has already been trained as a teacher. The teacher generates labels that a student policy consumes~\citep{hinton2015distilling}. 

Recent work has started using distillation for other applications than reducing model size. \cite{dacol2026opted} first trained a model with RL in vector-space. This allowed them to achieve super-human performance at low computational cost~\citep{cusumanotowner2025gigaflow}. Afterwards, \cite{dacol2026opted} distilled an end-to-end policy with vision encoder from the vector-space policy, achieving high performance at significantly less compute.

In our paper, in the context of autonomous driving, we distill by driving agents through a closed-loop simulation. At each step, the teacher generates a control, that the student uses as a label for its learning update. We distinguish between behavior cloning (BC)~\citep{bain1995behavioral-cloning}, where the teacher drives all agents at all times, and DAgger~\citep{ross2011dagger}, where over the course of the training, students are chosen to drive agents with increasing probability $1 - p_\mathrm{teacher}$. DAgger starts with BC, has a transition phase from $n_{1,\mathrm{DAgger}}$ to $n_{2,\mathrm{DAgger}}$, and finally lets all agents be driven by students. 
Intuitively, BC faces the issue of imitation learning that students don't learn to react to states that the teacher never visits, while DAgger teaches students to drive on their own, similar to RL. The distillation training loss is defined as
\begin{equation*}
    L_{u_0}(\theta) = \frac{1}{2B} \sum_{b=1}^{B} \left\| \frac{u_{0,b}(\theta) - u_b^\ast}{\tfrac12 (u_\mathrm{max} - u_\mathrm{min})} \right\|^2,
\end{equation*}
where $B$ is the batch size and $\theta$ are the network parameters. The teacher labels are $u_{b}^\ast$, the student prediction is $u_{0,b}(\theta)$, and $u_\mathrm{max}, u_\mathrm{min}$ are the upper and lower control limits.

\subsection{Model Predictive Control}
Model Predictive Control (MPC) uses known agent dynamics to forecast over a horizon $T$, how its control inputs $u_{0:T-1} \in \mathbb R^{m T}$ affect its future states $x_{0:T-1} \in \mathbb R^{n T}$. Only the first control $u_0$ is applied, and the problem is re-solved at the next step. We refer to the predicted timesteps $k = 0,\dots,T-1$ as the MPC's nodes. To track target states $\bar x_k$, we formulate a quadratic cost function
\begin{equation*}
    J(x_{0:T-1}, u_{0:T-1}) = \sum_{k=0}^{T-1} \tfrac12 (x_k - \bar x_k)^\top Q_k (x_k - \bar x_k) + \tfrac12 u_k^\top R_k u_k,
\end{equation*}
where $Q_k$ and $R_k$ are the state tracking and input regularization costs, respectively. By defining $\tau_k = [x_k;\; u_k] \in \mathbb R^{m+n}$, $\Xi_k = \mathrm{diag}(Q_k,\; R_k)$, and $\bar \tau_k = [\bar x_k;\; 0]$, the cost function simplifies to 
\begin{equation*}
    J(\tau_k) = \sum_{k=0}^{T-1} \tfrac12 (\tau_k - \bar \tau_k)^\top \Xi_k (\tau_k - \bar \tau_k).
\end{equation*}

We use the classic kinematic bicycle model (Appendix~\ref{app:dynamics}) to describe agent dynamics in our autonomous driving simulation. All positions and headings are given in the ego frame at the current step. The state $x_k$ is defined as $x_k = [p_{xk};\; p_{yk};\; \psi_k;\; v_k;\; \delta_k]$, where $p_{x}$ is the longitudinal progress, $p_y$ is the lateral progress, $v$ denotes the velocity, and $\delta$ and $\psi$ are the car's steering and yaw angles. The MPC is initialized at the current measured state $\hat x_0$ and follows control inputs $u_k = [a_k;\; \omega_k]$, where $a_k$ is the acceleration and $\omega_k$ is the steering speed.

The assumed kinematic bicycle model dynamics are non-linear, making this optimization problem nonconvex and difficult to solve~\citep[Sec.~4.2]{boyd2004convex}. As a remedy, we use the common approach of sequential quadratic programming, which linearizes the dynamics around the predicted states of an old solution, solves a Quadratic Program (QP), and repeats this process for $N_{\mathrm{SQP}}$ iterations. A QP consists of a quadratic term, linear dynamics, and linear constraints. Considering the dynamics $f(x,u): \mathbb R^{n} \times \mathbb R^m \rightarrow \mathbb R^n$, the diagonal cost terms $q_k \in \mathbb R^{n+m}$, and linear input constraints $u_\mathrm{min},\; u_\mathrm{max} \in \mathbb R^m$, our MPC optimization problem is defined as
\begin{equation}\label{eq:unconstrained-mpc}
    \begin{aligned}
        \min_{x_{0:T-1},\,u_{0:T-1}}\;
        &\sum_{k=0}^{T-1} \tfrac12 (\tau_k-\bar\tau_k)^\top \mathrm{diag}(q_k)\,(\tau_k-\bar\tau_k) \\
        \mathrm{s.t.} \qquad
        & x_{k+1} = f(x_k,u_k), && k = 0,\dots,T-2 \\
        & u_\mathrm{min} \le u_k \le u_\mathrm{max}, && k = 0,\dots,T-1 \\
        & x_0 = \hat x_0.
    \end{aligned}
\end{equation}

Besides the MPC's ability to optimize its control inputs by planning for a longer horizon, it is able to enforce input and state constraints. Equation~\eqref{eq:unconstrained-mpc} already enforces linear input constraints on $u_k$. In this work, we refer to this optimization problem as an \emph{unconstrained MPC}, because it has no state constraints on $x_k$. We discuss the MPC formulation with state constraints in Section~\ref{sec:constrained-mpc}.

\subsection{Differentiable MPC}\label{sec:diffmpc}
MPC assumes full knowledge of the system. This is often sufficiently accurate for vehicle dynamics and control limits, yet it becomes challenging when choosing tracking targets or a cost function. \cite{amos2018differentiable} showed that an MPC can be differentiated through, using its KKT conditions and the implicit function theorem. This allows us to use differentiable MPC (DiffMPC) as an action head in end-to-end policies, where an MPC adapter converts the backbone's hidden states to costs $q_k$ and targets $\bar \tau$. %We discuss Section~\ref{sec:methods}. 

This approach requires many samples to train an end-to-end policy in simulation. \cite{hose2026warpmpc} introduce WarpMPC, which can solve large batches of optimization problems on the GPU. WarpMPC is capable of computing forward passes to solve our MPC, and backward passes to train our end-to-end policy. Note that while WarpMPC significantly reduces the computational cost of solving large MPC batches, it remains more expensive than a simple MLP forward pass. Detailed differences in training time can be found in Appendix~\ref{app:training}. 

We refer to the entire end-to-end policy with DiffMPC action head as a \emph{DiffMPC policy}. We use the unconstrained DiffMPC policy to show that DiffMPC policies match an MLP's expressiveness. The constrained DiffMPC policy, as discussed in Section~\ref{sec:constrained-mpc}, is the main finding of our work.

\section{Methods}\label{sec:methods}

\begin{figure}[t]
    \centering
    \includegraphics[width=\textwidth]{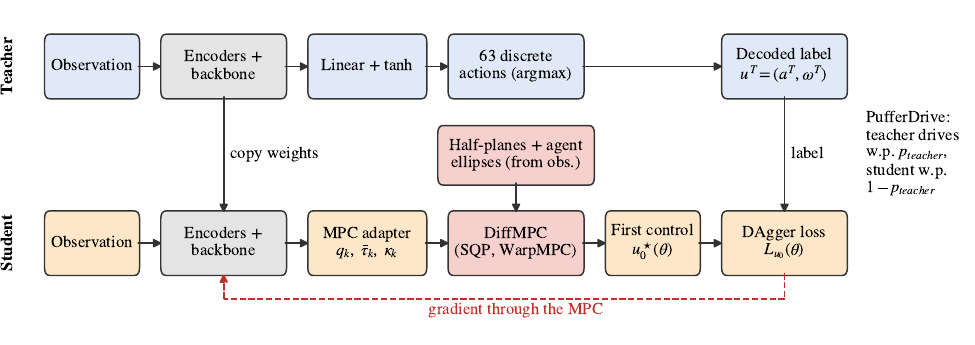}
    \vspace{-0.4in}
    \caption{Overview of our pipeline, including the backbone transfer and DAgger distillation. 
    }
    \label{fig:pipeline}
\end{figure}
Recall that our ultimate goal is to add verifiable constraints to a strong end-to-end driving policy without retraining it from scratch and without losing its performance. Our proposed pipeline to achieve this has three steps. First, we train an end-to-end teacher policy with RL in PufferDrive. Second, we transfer the teacher's encoder and backbone to the student, and replace the teacher's action head with an MPC adapter and DiffMPC. Third, we distill the new architecture from the teacher with DAgger. Figure~\ref{fig:pipeline} depicts a schematic overview of the second and third steps. 

We note that our DiffMPC policy is not intended to compete with the original end-to-end policy, but rather is a method to add constraints after training through an action head change and consecutive fine-tuning step. This keeps the training cost far below RL-only approaches. 
Section~\ref{sec:constraints} describes the constraints we use in our constrained DiffMPC formulation. Section~\ref{sec:constrained-mpc} details the constrained DiffMPC setup, followed by Section~\ref{sec:adapter}, which describes the MPC adapter.

\subsection{Constraints}\label{sec:constraints}
Agents in PufferDrive have access to information on roads, boundaries, and other agents. Such information can be provided through separate computer vision~\citep{li2022bevformer} in real-world settings. Utilizing the high accuracy of dedicated computer vision adds redundancy during operation. 
Next, we present how road boundary constraints and ellipse constraints of other agents are computed.

\subsubsection{Road boundary constraints}

\begin{wrapfigure}[13]{r}{0.36\textwidth}
    \vspace{-2em}
    \centering
    \includegraphics[width=0.32\textwidth]{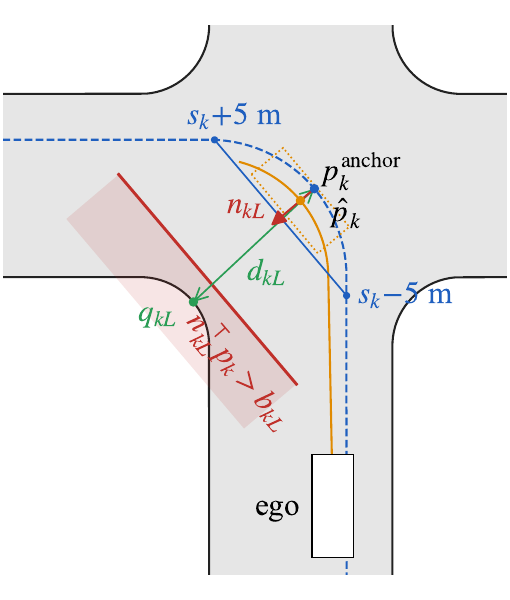}
    \caption{Left half-plane of node $k$ in a left turn. 
    }
    \label{fig:route-boundary}
    \vspace{-3em}
\end{wrapfigure}
Road boundary constraints are half-planes that are meant to stop the car from leaving the road. In four steps, we compute a left and right half-plane $h\in \{L,\; R\}$ for each node $k \ge 1$ that the MPC predicts.

\emph{Anchor.} We project the reference position $\hat p_k$, i.e., the previous MPC plan shifted by one step, onto the route. This gives the anchor $p_k^\mathrm{anchor}$ and its arc length $s_k$ along the route. 

\emph{Direction.} The route tangent $t_k$ is the chord between the route points at $s_k \pm 5$\,m. With its left normal $\nu_k$, the plane normals are $n_{kL} = \nu_k$ and $n_{kR} = -\nu_k$.

\emph{Distance.} $d_{kh}$ is the lateral distance from the anchor to the nearest road edge on side $h$. An edge only counts if it is within $35^\circ$ of parallel to $t_k$, within $\pm 2$\,m along the route, and on side $h$ of $\hat p_k$. 

\emph{Footprint.} The car's body is kept inside the plane with $\mathrm{offset}_{kh} = \tfrac{l_e}{2}|n_{kh}^\top \hat t_k| + \tfrac{w_e}{2}|n_{kh}^\top \hat \nu_k| + 0.2$\,m, where $\hat t_k$ and $\hat \nu_k$ are the planned heading at node $k$ and its normal, and $l_e$ and $w_e$ are the ego's length and width.

With $b_{kh} = n_{kh}^\top p^{\text{anchor}}_k + d_{kh} - \text{offset}_{kh}$, the resulting constraints with slack are:
\begin{equation}\label{eq:boundary-constraint}
    n_{kh}^\top p_k - b_{kh} \le s_{kh}, \quad s_{kh}\ge 0, \quad k \ge 1,\ h \in \{L,R\}.
\end{equation}
The half-planes are convex in $p_k$ and rebuilt from the observation at every step. If the road is narrower than the car, i.e., the anchor itself violates a plane, that plane is dropped.

\subsubsection{Agent ellipse constraints}
To avoid hitting other agents, we use elliptical constraints. 
We take the $K$ nearest agents from the observation. Each agent $i$ is predicted at constant velocity along its heading $h_i = (\cos\psi_i,\ \sin\psi_i)$, i.e., $c_{ki} = c_{0i} + v_i h_i\, k\, dt$. Each agent gets a fixed ellipse through the corners of the combined box of the agent and the ego:
\begin{equation*}
    a_i = \sqrt2\cdot\tfrac12\,(l_i + l_e), \qquad b_i = \sqrt2\cdot\tfrac12\,(w_i + w_e).
\end{equation*}
This way, the ellipse covers the full footprint of both cars when their headings are aligned. For large heading differences, e.g., in crossing traffic, it covers them only approximately. If fewer than $K$ agents are observed, the empty slots get a unit circle 500\,m ahead that never constrains the plan.

The resulting constraints with slack are:
\begin{equation}\label{eq:ellipse-constraint}
    \sqrt{\Big(\tfrac{(p_k-c_{ki})^\top h_i}{a_i}\Big)^2 + \Big(\tfrac{h_i \times (p_k-c_{ki})}{b_i}\Big)^2 + 10^{-4}} + \xi_{ki} \ge 1, \quad \xi_{ki}\ge 0, \quad k\ge 1,\ i = 1,\ldots,K.
\end{equation}
The first term is the distance along the agent's heading and the second the distance across it, where $h \times d = h_x d_y - h_y d_x$ is the 2D cross product. 
The constraints are not convex, but the SQP linearizes them at every iteration, so each QP sees the tangent half-plane of the ellipse at the current plan.

\subsection{Constrained DiffMPC}\label{sec:constrained-mpc}

We extend the unconstrained MPC~\eqref{eq:unconstrained-mpc} with the road boundary constraints and agent ellipses from~\eqref{eq:boundary-constraint} and~\eqref{eq:ellipse-constraint}, respectively. During operation, agents face scenarios where they cannot avoid violating constraints. This can happen before a crash or because of incorrectly set constraints. To avoid losing such gradients during training, we use slack variables. Slack variables become positive once a constraint is violated and must be higher penalized than the remaining cost term at a node. In the constrained MPC formulation, we use a normalized cost $\tilde q_k$, such that $1^\top \tilde q_k = 1$. The quadratic cost-term becomes $\tfrac12$ if all values in $\tau$ are at their per-unit maximum of $1$. A linear slack penalty is exact, i.e., the MPC keeps the constraints whenever it can, if its price exceeds the constraints' Lagrange multipliers~\citep{kerrigan2000soft}. Fixing the sum of the tracking weights keeps these multipliers bounded, so a fixed slack price stays above them in practice. %{\color{red} Would be great to have a proof or citation on linear slack penalties.}

Normalizing the cost has the unwanted consequence that all nodes along the MPC's trajectory are weighted equally. This is an unwanted constraint that limits the network's freedom. Thus, we introduce the node weight $\kappa_k$, which scales tracking and slack costs equally to preserve their relative weight, but allows the network to choose different weights along the MPC horizon. 

The final form of our proposed constrained DiffMPC is as follows:
\begin{equation}\label{eq:constrained-mpc}
    \begin{aligned}
        &\min_{x_{0:T-1},\,u_{0:T-1},\,s,\,\xi}\;
        \sum_{k=0}^{T-1} \kappa_k\, \tfrac12\,(\tau_k-\bar\tau_k)^\top \mathrm{diag}(\tilde q_k)\,(\tau_k-\bar\tau_k)
        + \sum_{k=1}^{T-1} \kappa_k\, \Big[\sum_{h\in\{L,R\}} s_{kh} + \sum_{i=1}^{K} \xi_{ki}\Big] \\
        &\qquad\quad\;\begin{aligned}
        \mathrm{s.t.} \qquad\;
        & x_0 = \hat x_0, \qquad x_{k+1} = f(x_k,u_k), && k = 0,\dots,T-2 \\
        & u_\mathrm{min} \le u_k \le u_\mathrm{max}, && k = 0,\dots,T-1 \\
        &\text{road boundary constraints, see \eqref{eq:boundary-constraint}}, \\
        &\text{ellipse constraints, see \eqref{eq:ellipse-constraint}}.
        \end{aligned}
    \end{aligned}
\end{equation}

\subsection{MPC adapter}\label{sec:adapter}
The MPC adapter is a linear layer from the backbone's latent states to logits. With slight abuse of notation, we denote the logits for node $k$ as $z_{1:n_\mathrm{pred},k} \in \mathbb R^{n_\mathrm{pred}T}$, where $n_\mathrm{pred}$ denotes the DiffMPC policy's number of learned values per node. Both policies have four target logits $z_{1:4,k}$ and six cost term logits $z_{5:10,k}$. The constrained DiffMPC policy has an additional logit $z_{11,k}$ for the node weight.\looseness=-1 

\emph{For both DiffMPC policies,} the longitudinal weight $q_{kx}$ is always zero, because progress is set by the speed target, and a position target along the lane would conflict with it. Thus, we also set the longitudinal progress target $\bar p_x = 0$. The remaining targets are set in per-unit scale through the logits $z_{1:4,k}$:
\begin{equation*}
\end{equation*}

\emph{For the unconstrained DiffMPC policy,} the state penalties are scaled with a sigmoid to $[0,10]$, and the control penalties to $[0.1,10]$. It follows the weight function:
\begin{equation}\label{eq:unconstrained-weights}
    q_k = \big(0,\; 10\,\sigma(z_{5:8,k}),\; 0.1 + 9.9\,\sigma(z_{9:10,k})\big).
\end{equation}

\emph{For the constrained DiffMPC policy,} the weight logits belong to the terms $j \in \{p_y, \psi, v, \delta, a, \omega\}$, and $\tilde q_{kj}$ denotes the entry of $\tilde q_k$ that weighs term $j$. The entry for $p_x$ is always zero. We turn them into normalized weights with a softmax and keep a floor $\epsilon = 0.01$ on each control term, so that the controls are always regularized:
\begin{equation*}
    \tilde q_{kj} = (1 - 2\epsilon)\,\mathrm{softmax}(z_{5:10,k})_j + \epsilon\,\mathbf{1}[j \in \{a, \omega\}].
\end{equation*}
The node weight is centered over the horizon, so that equal logits give $\kappa_k = 1$:
\begin{equation*}
    \kappa_k = 1 + 0.75\tanh\!\big(z_{11,k} - \bar z_{11}\big), \qquad \bar z_{11} = \tfrac1T \textstyle\sum_{k'=0}^{T-1} z_{11,k'}.
\end{equation*}

\section{Experiments}

\subsection{Setup}\label{sec:setup}

\begin{wrapfigure}[12]{r}{0.42\textwidth}
    \vspace{-1.5em}
    \centering
    \includegraphics[width=0.4\textwidth]{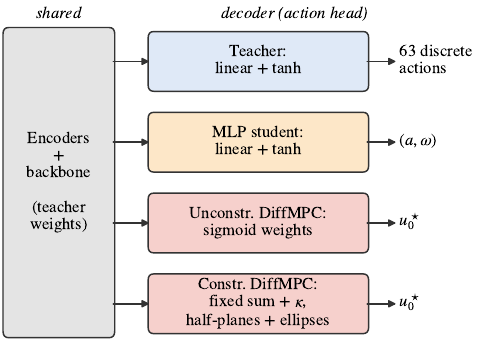}
    \caption{All policies share the teacher's encoder and backbone.}
    \label{fig:students}
    \vspace{-1em}
\end{wrapfigure}
\emph{Teachers.} We demonstrate our pipeline on two teachers, a smaller 4x512 and a larger 4x1024 MLP policy. Both are trained with PPO in a staged curriculum on CARLA maps and then fine-tuned on a mix of CARLA and nuPlan maps, for 1.1B and 7.9B steps, respectively. The training stages and hyperparameters are given in Appendix~\ref{app:teacher}.

\emph{Students and baselines.} For each teacher size, we compare three students, shown in Figure~\ref{fig:students}. All students reuse the encoder and backbone of their teacher and only differ in their decoder. The MLP student replaces the teacher's discrete actions with a continuous linear head; this student helps us assess the cost of the distillation itself. The unconstrained DiffMPC student uses the MPC head without constraints and with the sigmoid weights in~\eqref{eq:unconstrained-weights}. This student shows whether the MPC head is expressive enough to match the teacher. Finally, the constrained DiffMPC student is our main method, with the fixed-sum weights, the node weights $\kappa_k$, the road boundary constraints, and the agent ellipses. 

\emph{MPC settings.} The MPC plans over $T=10$ nodes with sampling time $dt=0.1$\,s, i.e., one second ahead. WarpMPC runs five SQP iterations with 25 ADMM iterations each and $\rho=3$. We use $K=4$ agent ellipses and control limits $- u_\mathrm{min} = u_\mathrm{max} = [4;\; 0.75]$, which are normalized to per-unit scale in the MPC formulation. 

\emph{Training and model selection.} All students are distilled with DAgger on CARLA maps, using Adam with learning rate $3\cdot10^{-4}$ and batches of 1024 labels. Every 10M labels, we evaluate the student on 250 CARLA validation scenes. Training stops after two consecutive evaluations without improvement, where an improvement is a lower sum of collision and off-road rates or at least 0.03 more goals per vehicle, and after at most 100M labels. We select the checkpoint with the lowest sum of collision and off-road rates on the validation set. Only this checkpoint is evaluated on the test sets. We train nine seeds per student and teacher size on one RTX 5090. 

\emph{Evaluation.} We test on 1000 CARLA scenes and 3645 nuPlan scenes. Additionally, the Waymo Open Dataset (WOD)~\citep{ettinger2021womd} is not used to train the teachers or the students, so we use it as an out-of-distribution (OOD) test. We evaluate on 1000 Waymo validation scenes with $9\,$s episodes. We report the number of goals reached per vehicle, and the rates of collisions and off-road driving as mean $\pm$ standard error over the nine seeds. We compare students with a two-sided Welch $t$-test, Holm-corrected over the six comparisons per policy size. For simplicity, we refer to the sum of collisions and off-road rates as failure rates.

\subsection{Results}\label{sec:results}
Table~\ref{tab:main} shows the test results of all policies. All policies of the same size reach a similar number of goals, so we compare them by their combined collision and off-road (failure) rates. We start by discussing the CARLA and nuPlan results (denoted as CARLA / nuPlan). 

\emph{Teachers.} The small and large teacher show large performance differences. The large teacher's failures drop to $21\%$ / $22\%$ of the small teacher's failures, while achieving $2.4\times$ / $1.4\times$ the number of goals. This provides us with two cases: a weak policy that needs constraints for safety, and a capable policy that needs constraints for verifiability.

\begin{table}[t]
    \centering
    \small
    \begin{tabular}{llcccccc}
        \hline
        & & \multicolumn{2}{c}{CARLA} & \multicolumn{2}{c}{nuPlan} & \multicolumn{2}{c}{Waymo (OOD)} \\
        Size & Policy & Goals & Failures & Goals & Failures & Goals & Failures \\
        \hline
        4x512 & Teacher & 1.67 & .245 & 0.77 & .181 & 0.40 & .298 \\
              & MLP & 1.66 & .226$\pm$.004 & 0.80 & .205$\pm$.006 & 0.36 & .299$\pm$.007 \\
              & Unconstr.\ DiffMPC & 1.68 & .224$\pm$.003 & 0.82 & .220$\pm$.005 & 0.35 & .280$\pm$.008 \\
              & Constr.\ DiffMPC & 1.68 & \textbf{.180}$\pm$.003 & 0.81 & \textbf{.180}$\pm$.006 & 0.38 & \textbf{.261}$\pm$.008 \\
        \hline
        4x1024 & Teacher & 3.95 & .052 & 1.05 & .039 & 0.59 & .059 \\
               & MLP & 3.95 & \textbf{.050}$\pm$.001 & 1.05 & .040$\pm$.001 & 0.60 & .059$\pm$.002 \\
               & Unconstr.\ DiffMPC & 3.94 & .052$\pm$.002 & 1.06 & .044$\pm$.003 & 0.61 & .059$\pm$.002 \\
               & Constr.\ DiffMPC & 3.91 & .053$\pm$.001 & 1.05 & \textbf{.035}$\pm$.001 & 0.59 & \textbf{.046}$\pm$.002 \\
        \hline
    \end{tabular}
    \caption{Test results per vehicle: goals reached and failure rate (mean $\pm$ standard error over nine seeds; each teacher is a single policy). 
    }
    \label{tab:main}
    \vspace{-1.5em}
\end{table}

\emph{Distillation losses.} The MLP students stay close to their teachers. For the large policy, the MLP student reaches 0.050 / 0.040 against the teacher's 0.052 / 0.039 (CARLA / nuPlan). Thus, DAgger with the teacher's backbone loses almost no performance for the large policy. For the small policy, distillation costs some performance on nuPlan, where the MLP student reaches $0.205$ against the teacher's $0.181$.

\emph{DiffMPC expressiveness.} The unconstrained DiffMPC students perform similarly to the MLP students. For the large policy, they reach $0.052$ / $0.044$. For the small policy, they are on par on CARLA and slightly worse on nuPlan. Thus, we conclude that the MPC head is expressive enough to replace the teacher's action head. 

\emph{DiffMPC constraints.} We now turn to our main comparison between the teacher and the constrained student, which adds explicit constraints without losing performance. 
For the small policy, the constraints clearly help on CARLA by reducing failure rates from the teacher's $0.245$ to the student's $0.180\pm 0.003$. On nuPlan the constrained DiffMPC stays on par with the teacher's failure rates, while the other students fail more often. For the large policy, the constrained DiffMPC student has the lowest failure rates on nuPlan and matches the teacher on CARLA within one standard error. Generally, the constrained DiffMPC student successfully keeps up with the teacher. Across both sizes and all three test sets, it is never significantly worse than the MLP or unconstrained students.

We believe that whether the constraints help depends on the quality of the constraints compared to the teacher. The small teacher makes more mistakes than the constraints allow, so the constraints correct it. The large teacher already drives well, so the constraints rarely add information.

\emph{Generalization.} We next use the WOD data to measure the policies' OOD performance, and therefore, their ability to generalize. For both policy sizes, the constrained students fail less often than their teachers and MLP students, mostly through less off-road driving. For the large policies, the off-road rate drops from $0.035$ to $0.025$, which shows the effectiveness of the road boundary constraints. Their effectiveness suggests that the constraints generalize better than the learned policy. While our constraints use ground-truth map data, they can be seen as an independent measurement, that increases system robustness in scenarios that are new to the end-to-end policy.

\emph{Training cost.} All students reach their best validation checkpoint within 10M to 60M labels, compared to the 1.1B and 7.9B PPO steps of the small and large teacher, respectively. Distilling the constrained student takes $2.8$--$4.7\,$h and $2.0$--$3.3\,$h on one GPU for the small and large policy, on top of the teachers' $1\,$h and $11.4\,$h of training. For the large policy, this adds at most $30\%$ to the teacher's training time. For the small policy, distillation takes longer than training the teacher, because each MPC label costs about eight to ten times an MLP label, while the small teacher converged quickly. Appendix~\ref{app:training} details the training times and validation curves, and Appendix~\ref{app:ablation} shows why the constrained DiffMPC needs the fixed-sum weights and the node weights $\kappa_k$.

\section{Conclusion and future work}

This work presented a new pipeline that helps adding redundancy to safety critical systems with the help of DiffMPC. Besides providing a novel method for training a DiffMPC policy, we change the perspective on it from a competing architecture to a new post-training step for end-to-end MLP policies. Results clearly show that (1) unconstrained DiffMPC policies are as expressive as MLP policies (2) constrained DiffMPC adds redundancy with constraints, which leads to reduced failure rates for smaller, more failure-prone policies, and on the out-of-distribution WOD scenes. 

Future work should look into designing better constraints and verifying them on larger end-to-end models. Another intuitive extension is to use the new pipeline outside of vectorized simulation and showing its potential in applications beyond autonomous driving.

\subsection*{Generative AI Use}

The research questions, method, experimental design, and analysis of results are the authors' own. Generative AI tools wrote the code for the experiments, under the authors' specification and review. They were also used to support literature search and idea generation, and to assist with the writing of the manuscript. The authors verified all AI-generated code and text and take full responsibility for the content of this paper.

\bibliography{l4dc2027-sample}

@inproceedings{amos2018differentiable,
  title     = {Differentiable {MPC} for End-to-end Planning and Control},
  author    = {Amos, Brandon and Jimenez Rodriguez, Ivan Dario and Sacks, Jacob and Boots, Byron and Kolter, J. Zico},
  booktitle = {Advances in Neural Information Processing Systems},
  volume    = {31},
  year      = {2018}
}

@InProceedings{ross2011dagger,
  title = 	 {A Reduction of Imitation Learning and Structured Prediction to No-Regret Online Learning},
  author = 	 {Ross, Stephane and Gordon, Geoffrey and Bagnell, Drew},
  booktitle = 	 {Proceedings of the Fourteenth International Conference on Artificial Intelligence and Statistics},
  pages = 	 {627--635},
  year = 	 {2011},
  volume = 	 {15},
  series = 	 {Proceedings of Machine Learning Research},
  address = 	 {Fort Lauderdale, FL, USA},
  month = 	 {11--13 Apr},
  publisher =    {PMLR},
  url = 	 {https://proceedings.mlr.press/v15/ross11a.html},
}

@misc{hose2026warpmpc,
      title={WarpMPC: Large-Batch MPC on GPU via ADMM with Unrolled $LDL^\top$ Factorization}, 
      author={Henrik Hose and Se Hwan Jeon and Charles Khazoom and Sangbae Kim and Sebastian Trimpe},
      year={2026},
      eprint={2607.11603},
      archivePrefix={arXiv},
      primaryClass={cs.RO},
      url={https://arxiv.org/abs/2607.11603}, 
}

@misc{dacol2026opted,
      title={OPTED: On-Policy Fine-Tuning for End-to-End Driving using a Render-Free Teacher}, 
      author={Damiano Da Col and Maximilian Igl and Peter Karkus and Kashyap Chitta and Boris Ivanovic and Marco Pavone and Konrad Schindler and Christos Sakaridis},
      year={2026},
      eprint={2609.20756},
      archivePrefix={arXiv},
      primaryClass={cs.RO},
      url={https://arxiv.org/abs/2609.20756}, 
}

@InProceedings{rickenbach2025zipmpc,
  title = 	 {ZipMPC: Compressed Context-Dependent MPC Cost via Imitation Learning},
  author =       {Rickenbach, Rahel and Lahoud, Alan and Schaffernicht, Erik and Zeilinger, Melanie and Stork, Johannes A.},
  booktitle = 	 {Proceedings of The 9th Conference on Robot Learning},
  pages = 	 {3115--3136},
  year = 	 {2025},
  volume = 	 {305},
  series = 	 {Proceedings of Machine Learning Research},
  month = 	 {27--30 Sep},
  publisher =    {PMLR},
  url = 	 {https://proceedings.mlr.press/v305/rickenbach25a.html},
}

@article{romero2026acmpc,
  title = {Actor--Critic Model Predictive Control: Differentiable Optimization Meets Reinforcement Learning for Agile Flight},
  volume = {42},
  ISSN = {1941-0468},
  url = {http://dx.doi.org/10.1109/tro.2025.3644945},
  DOI = {10.1109/tro.2025.3644945},
  journal = {IEEE Transactions on Robotics},
  publisher = {Institute of Electrical and Electronics Engineers (IEEE)},
  author = {Romero, Angel and Aljalbout, Elie and Song, Yunlong and Scaramuzza, Davide},
  year = {2026},
  pages = {673--692}
}

@misc{wu2026acsmpc,
      title={AC-SMPC: Differentiable Actor--Critic MPC with State Constraints for Safe and Fast Robot Navigation}, 
      author={Chaoran Wu and Jiayu Fan and Qichen Tan and Yuyi Liu and Jin Wang},
      year={2026},
      eprint={2610.41477},
      archivePrefix={arXiv},
      primaryClass={cs.RO},
      url={https://arxiv.org/abs/2610.41477}, 
}

@misc{zhang2026smoothedturbompc,
      title={Efficient Learning of Differentiable MPC Policies on the GPU via Smoothed ADMM and First-Order Reinforcement Learning}, 
      author={Jianghan Zhang and Roland Schwan and Gabriel Bravo-Palacios and John Zhang and Thomas Lew and Brian Plancher},
      year={2026},
      eprint={2611.09937},
      archivePrefix={arXiv},
      primaryClass={cs.RO},
      url={https://arxiv.org/abs/2611.09937}, 
}

@article{papuc2026lmpg,
  title = {Strategizing at Speed: A Learned Model Predictive Game for Multi-Agent Drone Racing},
  volume = {11},
  ISSN = {2377-3774},
  url = {http://dx.doi.org/10.1109/lra.2026.3701568},
  DOI = {10.1109/lra.2026.3701568},
  number = {8},
  journal = {IEEE Robotics and Automation Letters},
  publisher = {Institute of Electrical and Electronics Engineers (IEEE)},
  author = {Papuc, Andrei-Carlo and Peters, Lasse and Sun, Sihao and Ferranti, Laura and Alonso-Mora, Javier},
  year = {2026},
  month = Aug,
  pages = {9056--9063}
}

@misc{zarrouki2026sgrl,
      title={Accelerating Reinforcement Learning via MPC Solver-Gradient Guidance for Weights-varying MPC}, 
      author={Baha Zarrouki and Arslan Thobani and Jasper Hoffmann and Mattia Piccinini and Rudolf Reiter and Felix Jahncke and Sébastien Gros and Davide Scaramuzza and Johannes Betz},
      year={2026},
      eprint={2609.01061},
      archivePrefix={arXiv},
      primaryClass={cs.RO},
      url={https://arxiv.org/abs/2609.01061}, 
}

@misc{rowe2026gigapixel,
      title={Scaling Self-Play for End-to-End Driving}, 
      author={Luke Rowe and Roger Girgis and Rodrigue de Schaetzen and Daphne Cornelisse and Alaap Grandhi and Felix Heide and Eugene Vinitsky and Christopher Pal and Liam Paull},
      year={2026},
      eprint={2606.19641},
      archivePrefix={arXiv},
      note={CoRL 2026},
      primaryClass={cs.RO},
      url={https://arxiv.org/abs/2606.19641}, 
}

@misc{trisovic2026vigormpc,
      title={VIGOR-MPC: Vision-Guided Objective Regulation MPC}, 
      author={Jelena Trisovic and Rahel Rickenbach and Marek Landert and Panagiotis Minos and Melanie N. Zeilinger},
      year={2026},
      eprint={2610.20333},
      archivePrefix={arXiv},
      primaryClass={cs.RO},
      url={https://arxiv.org/abs/2610.20333}, 
}

@InProceedings{gurumurthy2025deqmpc,
  title = 	 {DEQ-MPC : Deep Equilibrium Model Predictive Control},
  author =       {Gurumurthy, Swaminathan and Nguyen, Khai and Bishop, Arun L and Kolter, J Zico and Manchester, Zachary},
  booktitle = 	 {Proceedings of The 9th Conference on Robot Learning},
  pages = 	 {961--980},
  year = 	 {2025},
  volume = 	 {305},
  series = 	 {Proceedings of Machine Learning Research},
  month = 	 {27--30 Sep},
  publisher =    {PMLR},
  url = 	 {https://proceedings.mlr.press/v305/gurumurthy25a.html},
}

@misc{bravopalacios2026turbompc,
      title={TurboMPC: Fast, Scalable, and Differentiable Model Predictive Control on the GPU}, 
      author={Gabriel Bravo-Palacios and Jianghan Zhang and Zachary Pestrikov and Brian Plancher and Thomas Lew},
      year={2026},
      eprint={2606.24039},
      archivePrefix={arXiv},
      primaryClass={cs.RO},
      url={https://arxiv.org/abs/2606.24039}, 
}

@article{huang2024dipp,
  title = {Differentiable Integrated Motion Prediction and Planning With Learnable Cost Function for Autonomous Driving},
  volume = {35},
  ISSN = {2162-2388},
  url = {http://dx.doi.org/10.1109/tnnls.2023.3283542},
  DOI = {10.1109/tnnls.2023.3283542},
  number = {11},
  journal = {IEEE Transactions on Neural Networks and Learning Systems},
  publisher = {Institute of Electrical and Electronics Engineers (IEEE)},
  author = {Huang, Zhiyu and Liu, Haochen and Wu, Jingda and Lv, Chen},
  year = {2024},
  month = Nov,
  pages = {15222--15236}
}

@InProceedings{karkus2023diffstack,
  title = 	 {DiffStack: A Differentiable and Modular Control Stack for Autonomous Vehicles},
  author =       {Karkus, Peter and Ivanovic, Boris and Mannor, Shie and Pavone, Marco},
  booktitle = 	 {Proceedings of The 6th Conference on Robot Learning},
  pages = 	 {2170--2180},
  year = 	 {2023},
  volume = 	 {205},
  series = 	 {Proceedings of Machine Learning Research},
  month = 	 {14--18 Dec},
  publisher =    {PMLR},
  url = 	 {https://proceedings.mlr.press/v205/karkus23a.html},
}

@misc{cornelisse2026humanlike,
      title={Human-like autonomy emerges from self-play and a pinch of human data}, 
      author={Daphne Cornelisse and Julian Hunt and Zixu Zhang and Waël Doulazmi and Kevin Joseph and Jaime Fernández Fisac and Eugene Vinitsky},
      year={2026},
      eprint={2606.19370},
      archivePrefix={arXiv},
      note={CoRL 2026},
      primaryClass={cs.LG},
      url={https://arxiv.org/abs/2606.19370}, 
}

@misc{cusumanotowner2025gigaflow,
      title={Robust Autonomy Emerges from Self-Play}, 
      author={Marco Cusumano-Towner and David Hafner and Alex Hertzberg and Brody Huval and Aleksei Petrenko and Eugene Vinitsky and Erik Wijmans and Taylor Killian and Stuart Bowers and Ozan Sener and Philipp Krähenbühl and Vladlen Koltun},
      year={2025},
      eprint={2502.03349},
      archivePrefix={arXiv},
      primaryClass={cs.LG},
      url={https://arxiv.org/abs/2502.03349}, 
}

@misc{yin2026pictura,
      title={Pictura: Perspective-View Self-Play at Scale for Driving}, 
      author={Yuan Yin and Elias Ramzi and Marc Lafon and Valentin Charraut and Victor Bares and Yihong Xu and Éloi Zablocki and Alexandre Boulch and Thibault Buhet and Andrei Bursuc and Matthieu Cord},
      year={2026},
      eprint={2607.26005},
      archivePrefix={arXiv},
      primaryClass={cs.CV},
      url={https://arxiv.org/abs/2607.26005}, 
}

@misc{li2026worldengine,
      title={World Engine: Towards the Era of Post-Training for Autonomous Driving}, 
      author={Tianyu Li and Li Chen and Caojun Wang and Haochen Liu and Kashyap Chitta and Zhenjie Yang and Yuhang Lu and Naisheng Ye and Yihang Qiu and Yufei Wang and Luoxi Zou and Jiaxin Peng and Jin Pan and Zhaoyu Su and Andrei Bursuc and Shengbo Eben Li and Andreas Geiger and Peng Su and Hongyang Li},
      year={2026},
      eprint={2606.19836},
      archivePrefix={arXiv},
      primaryClass={cs.RO},
      url={https://arxiv.org/abs/2606.19836}, 
}

@misc{adabag2025diffmpcgpu,
      title={Differentiable Model Predictive Control on the GPU}, 
      author={Emre Adabag and Marcus Greiff and John Subosits and Thomas Lew},
      year={2025},
      eprint={2510.06179},
      archivePrefix={arXiv},
      primaryClass={math.OC},
      url={https://arxiv.org/abs/2510.06179}, 
}

@inproceedings{wabersich2018linear,
  title = {Linear Model Predictive Safety Certification for Learning-Based Control},
  url = {http://dx.doi.org/10.1109/cdc.2018.8619829},
  DOI = {10.1109/cdc.2018.8619829},
  booktitle = {2018 IEEE Conference on Decision and Control (CDC)},
  publisher = {IEEE},
  author = {Wabersich, Kim P. and Zeilinger, Melanie N.},
  year = {2018},
  month = Dec,
  pages = {7130--7135}
}

@inproceedings{dosovitskiy2017carla,
  title     = {{CARLA}: {An} Open Urban Driving Simulator},
  author    = {Dosovitskiy, Alexey and Ros, German and Codevilla, Felipe and Lopez, Antonio and Koltun, Vladlen},
  booktitle = {Proceedings of the 1st Annual Conference on Robot Learning},
  series    = {Proceedings of Machine Learning Research},
  volume    = {78},
  pages     = {1--16},
  year      = {2017},
  publisher = {PMLR},
  url       = {https://proceedings.mlr.press/v78/dosovitskiy17a.html}
}

@misc{caesar2021nuplan,
  title         = {{nuPlan}: A closed-loop {ML}-based planning benchmark for autonomous vehicles},
  author        = {Caesar, Holger and Kabzan, Juraj and Tan, Kok Seang and Fong, Whye Kit and Wolff, Eric and Lang, Alex and Fletcher, Luke and Beijbom, Oscar and Omari, Sammy},
  year          = {2021},
  eprint        = {2106.11810},
  archivePrefix = {arXiv},
  primaryClass  = {cs.CV},
  url           = {https://arxiv.org/abs/2106.11810}
}

@inproceedings{ettinger2021womd,
  title     = {Large Scale Interactive Motion Forecasting for Autonomous Driving: The {Waymo} Open Motion Dataset},
  author    = {Ettinger, Scott and Cheng, Shuyang and Caine, Benjamin and Liu, Chenxi and Zhao, Hang and Pradhan, Sabeek and Chai, Yuning and Sapp, Ben and Qi, Charles R. and Zhou, Yin and Yang, Zoey and Chouard, Aur{\'e}lien and Sun, Pei and Ngiam, Jiquan and Vasudevan, Vijay and McCauley, Alexander and Shlens, Jonathon and Anguelov, Dragomir},
  booktitle = {Proceedings of the IEEE/CVF International Conference on Computer Vision (ICCV)},
  pages     = {9690--9699},
  year      = {2021},
  doi       = {10.1109/ICCV48922.2021.00957}
}

@article{tang2025drlrobotics,
  title   = {Deep Reinforcement Learning for Robotics: A Survey of Real-World Successes},
  author  = {Tang, Chen and Abbatematteo, Ben and Hu, Jiaheng and Chandra, Rohan and Mart{\'\i}n-Mart{\'\i}n, Roberto and Stone, Peter},
  journal = {Annual Review of Control, Robotics, and Autonomous Systems},
  volume  = {8},
  year    = {2025}
}

@article{chen2024e2esurvey,
  title   = {End-to-End Autonomous Driving: Challenges and Frontiers},
  author  = {Chen, Li and Wu, Penghao and Chitta, Kashyap and Jaeger, Bernhard and Geiger, Andreas and Li, Hongyang},
  journal = {IEEE Transactions on Pattern Analysis and Machine Intelligence},
  volume  = {46},
  number  = {12},
  pages   = {10164--10183},
  year    = {2024}
}

@article{lemero2022ilsurvey,
  title   = {A Survey on Imitation Learning Techniques for End-to-End Autonomous Vehicles},
  author  = {Le Mero, Luc and Yi, Dewei and Dianati, Mehrdad and Mouzakitis, Alexandros},
  journal = {IEEE Transactions on Intelligent Transportation Systems},
  volume  = {23},
  number  = {9},
  pages   = {14128--14147},
  year    = {2022}
}

@article{kiran2022drlad,
  title   = {Deep Reinforcement Learning for Autonomous Driving: A Survey},
  author  = {Kiran, B. Ravi and Sobh, Ibrahim and Talpaert, Victor and Mannion, Patrick and Al Sallab, Ahmad A. and Yogamani, Senthil and P{\'e}rez, Patrick},
  journal = {IEEE Transactions on Intelligent Transportation Systems},
  volume  = {23},
  number  = {6},
  pages   = {4909--4926},
  year    = {2022},
}

@inproceedings{lu2023imitation,
  title     = {Imitation Is Not Enough: Robustifying Imitation with Reinforcement Learning for Challenging Driving Scenarios},
  author    = {Lu, Yiren and Fu, Justin and Tucker, George and Pan, Xinlei and Bronstein, Eli and Roelofs, Rebecca and Sapp, Benjamin and White, Brandyn and Faust, Aleksandra and Whiteson, Shimon and Anguelov, Dragomir and Levine, Sergey},
  booktitle = {IEEE/RSJ International Conference on Intelligent Robots and Systems (IROS)},
  pages     = {7553--7560},
  year      = {2023}
}

@article{kaufmann2023champion,
  title   = {Champion-level drone racing using deep reinforcement learning},
  author  = {Kaufmann, Elia and Bauersfeld, Leonard and Loquercio, Antonio and M{\"u}ller, Matthias and Koltun, Vladlen and Scaramuzza, Davide},
  journal = {Nature},
  volume  = {620},
  number  = {7976},
  pages   = {982--987},
  year    = {2023}
}

@incollection{bain1995behavioral-cloning,
  title     = {A Framework for Behavioural Cloning},
  author    = {Bain, Michael and Sammut, Claude},
  booktitle = {Machine Intelligence 15},
  editor    = {Furukawa, K. and Michie, D. and Muggleton, S.},
  publisher = {Oxford University Press},
  pages     = {103--129},
  year      = {1995} 
}

@article{hewing2020learning,
  title   = {Learning-Based Model Predictive Control: Toward Safe Learning in Control},
  author  = {Hewing, Lukas and Wabersich, Kim P. and Menner, Marcel and Zeilinger, Melanie N.},
  journal = {Annual Review of Control, Robotics, and Autonomous Systems},
  volume  = {3},
  pages   = {269--296},
  year    = {2020}
}

@article{reiter2025synthesis,
  title   = {Synthesis of Model Predictive Control and Reinforcement Learning: Survey and Classification},
  author  = {Reiter, Rudolf and Hoffmann, Jasper and Reinhardt, Dirk and Messerer, Florian and Baumg{\"a}rtner, Katrin and Sawant, Shambhuraj and Boedecker, Joschka and Diehl, Moritz and Gros, Sebastien},
  journal = {Annual Reviews in Control},
  year    = {2025}  
}

@article{schwarting2018safe,
  title   = {Safe Nonlinear Trajectory Generation for Parallel Autonomy With a Dynamic Vehicle Model},
  author  = {Schwarting, Wilko and Alonso-Mora, Javier and Paull, Liam and Karaman, Sertac and Rus, Daniela},
  journal = {IEEE Transactions on Intelligent Transportation Systems},
  volume  = {19},
  number  = {9},
  pages   = {2994--3008},
  year    = {2018},
  doi     = {10.1109/TITS.2017.2771351}
}

@article{liniger2015optimization,
  title   = {Optimization-based autonomous racing of 1:43 scale {RC} cars},
  author  = {Liniger, Alexander and Domahidi, Alexander and Morari, Manfred},
  journal = {Optimal Control Applications and Methods},
  volume  = {36},
  number  = {5},
  pages   = {628--647},
  year    = {2015},
  doi     = {10.1002/oca.2123}
}

@article{brunke2022safe,
  title   = {Safe Learning in Robotics: From Learning-Based Control to Safe Reinforcement Learning},
  author  = {Brunke, Lukas and Greeff, Melissa and Hall, Adam W. and Yuan, Zhaocong and Zhou, Siqi and Panerati, Jacopo and Schoellig, Angela P.},
  journal = {Annual Review of Control, Robotics, and Autonomous Systems},
  volume  = {5},
  number  = {1},
  pages   = {411--444},
  year    = {2022},
  doi     = {10.1146/annurev-control-042920-020211}
}

@article{yurtsever2020survey,
  title   = {A Survey of Autonomous Driving: Common Practices and Emerging Technologies},
  author  = {Yurtsever, Ekim and Lambert, Jacob and Carballo, Alexander and Takeda, Kazuya},
  journal = {IEEE Access},
  volume  = {8},
  pages   = {58443--58469},
  year    = {2020},
  doi     = {10.1109/ACCESS.2020.2983149}
}

@misc{hinton2015distilling,
  title         = {Distilling the Knowledge in a Neural Network},
  author        = {Hinton, Geoffrey and Vinyals, Oriol and Dean, Jeff},
  year          = {2015},
  note          = {NIPS 2014 Deep Learning Workshop},
  eprint        = {1503.02531}, archivePrefix = {arXiv}
}

@book{boyd2004convex,
  title     = {Convex Optimization},
  author    = {Boyd, Stephen and Vandenberghe, Lieven},
  publisher = {Cambridge University Press},
  address   = {Cambridge, UK},
  year      = {2004}
}

@inproceedings{li2022bevformer,
  title     = {{BEVFormer}: Learning Bird's-Eye-View Representation from Multi-Camera Images via Spatiotemporal Transformers},
  author    = {Li, Zhiqi and Wang, Wenhai and Li, Hongyang and Xie, Enze and Sima, Chonghao and Lu, Tong and Yu, Qiao and Dai, Jifeng},
  booktitle = {European Conference on Computer Vision (ECCV)},
  year      = {2022},
  eprint    = {2203.17270},
  archivePrefix = {arXiv}
}

@inproceedings{kerrigan2000soft,
  title     = {Soft Constraints and Exact Penalty Functions in Model Predictive Control},
  author    = {Kerrigan, Eric C. and Maciejowski, Jan M.},
  booktitle = {Proceedings of the UKACC International Conference on Control},
  address   = {Cambridge, UK},
  year      = {2000}
}

\appendix

\section{Classic kinematic bicycle model}\label{app:dynamics}
PufferDrive has multiple options to model the dynamics of vehicles. To keep the dynamics compatible with the discrete state-space system of an MPC, we use the classic kinematic bicycle model~\citep{cornelisse2026humanlike}. 
It is provided in PufferDrive and models the control inputs as acceleration along the agent's driving direction, and steering speed. A dynamic update is defined as follows:
\begin{equation}\label{eq:classic-kinematic-bicycle}
    \begin{aligned}
        \delta' &= \mathrm{clip}(\delta + \omega\,dt,\ \pm 0.667), \qquad v' = \mathrm{clip}(v + a\,dt,\ \pm 40),\\
        \beta &= \arctan\!\big(\tfrac12 \tan\delta'\big),\\
        p_x' &= p_x + v'\cos(\psi+\beta)\,dt, \qquad p_y' = p_y + v'\sin(\psi+\beta)\,dt,\\
        \psi' &= \psi + \frac{v'\cos\beta\,\tan\delta'}{\ell_\mathrm{wb}}\,dt.
    \end{aligned}
\end{equation}
Longitudinal progress along ego's driving direction is $p_x$. Its velocity is $v$. The lateral progress normal to ego's driving direction is $p_y$. The car's steering and yaw angles are $\delta$ and $\psi$, respectively. Additionally, the car uses a slip angle $\beta$. The control inputs are the acceleration $a$ and steering speed $\omega$. We denote the car's wheelbase with $\ell_\mathrm{wb} = 0.6\,l_e$, where $l_e$ is the car's length, and the simulation sampling time as $dt$. 

\section{Teacher training}\label{app:teacher}

\subsection{Training stages}
Both teachers are trained from scratch with PPO in a staged curriculum. Table~\ref{tab:teacher-stages} lists the stages. The 4x1024 teacher adds seven more CARLA stages with the same settings (2.3B CARLA steps in total), and then trains for 5.6B steps on the mix of CARLA and nuPlan maps. In the mixed stages, 3 workers run CARLA maps and 7 workers run nuPlan maps. The small teacher took about $1\,$h to train, whereas the large teacher took about $11.4\,$h. 

\begin{table}[h]
    \centering
    \begin{tabular}{lll}
        \hline
        Stage & Change & Steps (4x512 / 4x1024) \\
        \hline
        Pretrain & from scratch, short episodes, collision 2, comfort 0 & 300M / 300M \\
        Comfort & comfort auto & 200M / 200M \\
        Collision & collision 2 $\rightarrow$ 4 & 200M / 200M \\
        Long episodes & episode length 128 $\rightarrow$ 1280 & 200M / 1.6B \\
        Mixed fine-tune & 3 CARLA + 7 nuPlan workers & 200M / 5.6B \\
        \hline
        Total & & 1.1B / 7.9B \\
        \hline
    \end{tabular}
    \caption{Teacher training stages.}
    \label{tab:teacher-stages}
\end{table}

\subsection{Architecture}
The 4x512 teacher has a shared actor-critic network with a 64-wide encoder and no LayerNorm. The 4x1024 teacher has a separate actor and critic, a 128-wide encoder, and LayerNorm. Both use 63 discrete actions. The observation covers the ego state, the 16 nearest partners within 100\,m, 80 lane and 80 road boundary slots, the 4 nearest stop lines, and the goal and context.

\subsection{PPO hyperparameters}
\begin{table}[h]
    \centering
    \begin{tabular}{ll}
        \hline
        Optimizer & AdamW, learning rate $5\cdot10^{-4}$ with annealing \\
        Discount $\gamma$, GAE $\lambda$ & 0.999, 0.95 \\
        Clip coefficient & 0.2 \\
        Entropy coefficient & 0.01 \\
        Value function coefficient & 0.5 \\
        Gradient norm clip & 0.5 \\
        Minibatch size & 65{,}536 \\
        BPTT horizon & 128 \\
        Rollout & 10 workers $\times$ 1024 agents \\
        \hline
    \end{tabular}
    \caption{PPO hyperparameters of both teachers.}
    \label{tab:teacher-ppo}
\end{table}

\subsection{Reward weights}
The collision penalty is raised from 2 to 4 and the comfort penalty from 0 to auto during the curriculum. The off-road penalty is 1.5 and the reverse penalty 0.05. All other terms, such as goal, stop line, and alignment, keep PufferDrive's defaults.

\section{Student training times and validation curves}\label{app:training}
All students are trained on one RTX 5090.

Table~\ref{tab:timing} shows the training speed of the students, each measured alone on an otherwise idle GPU. The MPC head makes a label 5 to 10 times more expensive than the MLP head. Nearly all of this cost comes from solving the MPC during the rollouts, so the backbone size barely matters. Still, a full run of the large constrained student takes only 2.0 to 3.3\,h, including the validation evaluations.

\begin{table}[h]
    \centering
    \small
    \begin{tabular}{llccccc}
        \hline
        Size & Student & Labels/s & Min/10M & Val.\ (min) & Cost/label & Full run \\
        \hline
        4x512 & MLP & 63{,}650 & 2.6 & 0.9 & 1.0$\times$ & 15--22\,min \\
              & Unconstr.\ DiffMPC & 11{,}330 & 14.7 & 7.2 & 5.6$\times$ & 2.0\,h \\
              & Constr.\ DiffMPC & 6{,}630 & 25.1 & 13.1 & 9.6$\times$ & 2.8--4.7\,h \\
        \hline
        4x1024 & MLP & 57{,}200 & 2.9 & 1.0 & 1.0$\times$ & 13--17\,min \\
               & Unconstr.\ DiffMPC & 11{,}900 & 14.0 & 7.4 & 4.8$\times$ & 1.9--3.0\,h \\
               & Constr.\ DiffMPC & 7{,}180 & 23.2 & 13.2 & 8.0$\times$ & 2.0--3.3\,h \\
        \hline
    \end{tabular}
    \caption{Training speed of the students on one RTX 5090. A full run includes the validation evaluations and is computed from the times per 10M labels and the labels at which seeds 1--3 stopped (Tables~\ref{tab:val-1024} and~\ref{tab:val-512}).
    }
    \label{tab:timing}
\end{table}

Tables~\ref{tab:val-1024} and~\ref{tab:val-512} show the failure rate on the 250 CARLA validation scenes every 10M labels. The selected checkpoints are in bold. Most runs reach their best checkpoint early, within 10M to 30M labels. The DAgger ramp starts at $n_{1,\mathrm{DAgger}} = 4M$ and ends at $n_{2,\mathrm{DAgger}} = 12M$ labels. Training stops after two evaluations in a row without improvement, i.e., without a lower failure rate or at least 0.03 more goals per vehicle. The number of goals stays at 3.9--4.0 for the large and 1.6--1.8 for the small students in all evaluations. The only exception is the third seed of the large unconstrained student, which collapses at 50M labels (failure rate 0.205, 3.13 goals). This is a closed-loop instability after its selected checkpoint at 30M. % TODO: add seed 4 once finished

\begin{table}[h]
    \centering
    \small
    \begin{tabular}{lcccccccc}
        \hline
        4x1024 & 10M & 20M & 30M & 40M & 50M & 60M & 70M & 80M \\
        \hline
        Constr.\ DiffMPC s1 & \textbf{.049} & .053 & .056 & .066 & .060 & & & \\
        Constr.\ DiffMPC s2 & \textbf{.055} & .061 & .056 & & & & & \\
        Constr.\ DiffMPC s3 & \textbf{.056} & .062 & .061 & .057 & & & & \\
        Unconstr.\ DiffMPC s1 & .047 & .048 & .045 & .052 & \textbf{.041} & .046 & .050 & \\
        Unconstr.\ DiffMPC s2 & .051 & .065 & .061 & .050 & .051 & \textbf{.049} & .056 & .055 \\
        Unconstr.\ DiffMPC s3 & .055 & .060 & \textbf{.055} & .063 & .205 & & & \\
        MLP s1 & \textbf{.046} & .056 & .053 & & & & & \\
        MLP s2 & .062 & \textbf{.049} & .055 & .053 & & & & \\
        MLP s3 & .057 & \textbf{.047} & .062 & .055 & & & & \\
        \hline
    \end{tabular}
    \caption{Failure rate of the 4x1024 students (seeds 1--3) on the 250 CARLA validation scenes.}
    \label{tab:val-1024}
\end{table}

\begin{table}[h]
    \centering
    \small
    \begin{tabular}{lccccccc}
        \hline
        4x512 & 10M & 20M & 30M & 40M & 50M & 60M & 70M \\
        \hline
        Constr.\ DiffMPC s1 & .199 & \textbf{.184} & .205 & .202 & & & \\
        Constr.\ DiffMPC s2 & .206 & .224 & .202 & .224 & \textbf{.179} & .194 & .210 \\
        Constr.\ DiffMPC s3 & .192 & \textbf{.173} & .199 & .197 & .177 & & \\
        Unconstr.\ DiffMPC s1 & .236 & .243 & \textbf{.222} & .246 & .242 & & \\
        Unconstr.\ DiffMPC s2 & .257 & .217 & \textbf{.217} & .250 & .238 & & \\
        Unconstr.\ DiffMPC s3 & .251 & .249 & \textbf{.228} & .241 & .241 & & \\
        MLP s1 & \textbf{.225} & .226 & .231 & .227 & & & \\
        MLP s2 & .222 & \textbf{.218} & .252 & .240 & & & \\
        MLP s3 & .258 & .260 & .234 & \textbf{.225} & .236 & .232 & \\
        \hline
    \end{tabular}
    \caption{Failure rate of the 4x512 students (seeds 1--3) on the 250 CARLA validation scenes.}
    \label{tab:val-512}
\end{table}

\section{Constraint formulation}\label{app:ablation}
Table~\ref{tab:ablation} shows how the constrained DiffMPC developed, measured as the sum of collision and off-road rates on the 250 CARLA validation scenes. All rows are single 4x1024 runs, distilled with DAgger on CARLA maps, at their selected checkpoint. % All rows are 4x1024. TODO: add the teacher's validation score
Adding the road boundary constraints and agent ellipses to the unconstrained student only at evaluation strongly degrades its performance, so the constraints have to be present during training. Training with the constraints improves the student, but it remains well below the unconstrained student. In these first constrained runs, each slack variable has a fixed linear price of 100. With the sigmoid weights in~\eqref{eq:unconstrained-weights}, the student learns weights that sum up to about 825 per node, which outweighs the slack price and effectively softens the constraints. The large weights also slow down the ADMM. Capping each weight does not help either, as the predictor can still shrink all weights. Fixing the sum of the weights and pricing the slacks at the same sum, as in Section~\ref{sec:constrained-mpc}, keeps the constraints close to hard. Without the node weights, however, the student cannot follow the teacher where its line crosses a plane, and off-road driving grows with training. Only the fixed-sum weights with the node weights $\kappa_k$ bring the constrained student close to the unconstrained one.

\begin{table}[h]
    \centering
    \begin{tabular}{lc}
        \hline
        Student & Collision + off-road \\
        \hline
        Unconstrained DiffMPC & 0.045 \\
        \quad + road boundary constraints at evaluation only & 0.135 \\
        \quad + road boundary constraints and agent ellipses at evaluation only & 0.307 \\
        Trained with constraints, sigmoid weights & 0.110 \\
        Trained with constraints, capped sigmoid weights & 0.326 \\
        Trained with constraints, fixed-sum weights & 0.158 \\
        Trained with constraints, fixed-sum weights + $\kappa_k$ & 0.050 \\
        \hline
    \end{tabular}
    \caption{Constraint formulation variants of the 4x1024 student on the 250 CARLA validation scenes.}
    \label{tab:ablation}
\end{table}

\end{document}